\documentclass[10pt,twocolumn,letterpaper]{article}

\usepackage[pagenumbers]{cvpr} 

\usepackage[dvipsnames]{xcolor}

\definecolor{cvprblue}{rgb}{0.21,0.49,0.74}
\usepackage[pagebackref,breaklinks,colorlinks,citecolor=cvprblue]{hyperref}
\usepackage{multirow}
\usepackage{graphicx}
\usepackage{caption}

\def\paperID{12007} 
\def\confName{CVPR}
\def\confYear{2024}

\title{FG-MDM: Towards Zero-Shot Human Motion Generation via ChatGPT-Refined Descriptions}

\author{Xu Shi$^{1}$, Wei Yao$^{1}$, Chuanchen Luo$^{2}$, Junran Peng$^3$, Hongwen Zhang$^4$, Yunlian Sun$^{1}$\\
$^1$Nanjing University of Science and Technology 
\\ $^2$Shandong University\ \ $^3$Institute of Automation, Chinese Academy of Sciences
\\ $^4$Beijing Normal University
\\\href{https://sx0207.github.io/fg-mdm}{https://sx0207.github.io/fg-mdm}
}

\begin{document}

\twocolumn[{
\maketitle
\begin{center}
    \captionsetup{type=figure}
    \vspace{-8mm}
    \includegraphics[width=1.\textwidth]{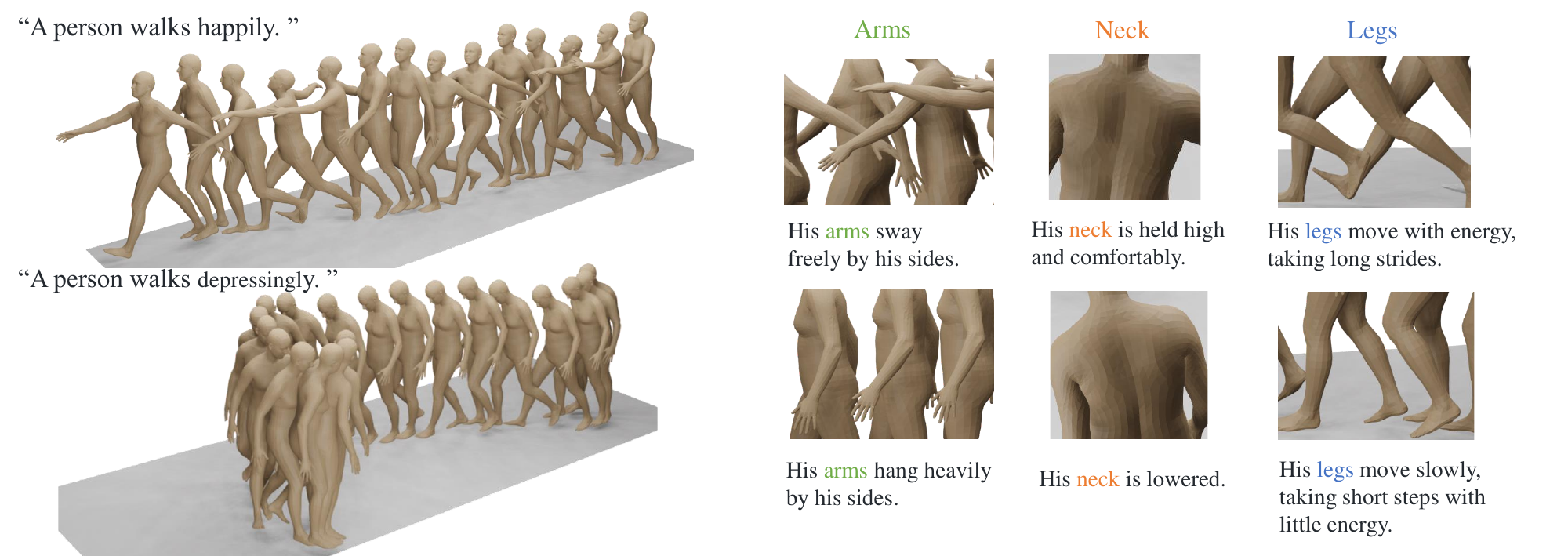}
    \vspace{-6mm}
    \captionof{figure}{FG-MDM can generate high-quality human motions in zero-shot settings by using fine-grained descriptions of different body parts.}
    \label{figshow}
\end{center}
}]

\maketitle


   
    

\begin{abstract}
Recently, significant progress has been made in text-based motion generation, enabling the generation of diverse and high-quality human motions that conform to textual descriptions. However, generating motions beyond the distribution of original datasets remains challenging, i.e., zero-shot generation. By adopting a divide-and-conquer strategy, we propose a new framework named Fine-Grained Human Motion Diffusion Model (FG-MDM) for zero-shot human motion generation. Specifically, we first parse previous vague textual annotations into fine-grained descriptions of different body parts by leveraging a large language model. We then use these fine-grained descriptions to guide a transformer-based diffusion model, which further adopts a design of part tokens. FG-MDM can generate human motions beyond the scope of original datasets owing to descriptions that are closer to motion essence. Our experimental results demonstrate the superiority of FG-MDM over previous methods in zero-shot settings. We will release our fine-grained textual annotations for HumanML3D and KIT.
\end{abstract}    
\section{Introduction}
\label{sec:intro}


Human motion generation is an important research topic in communities of both computer vision and computer graphics. It aims to simulate and generate realistic human movements using computers. With the advancement of technologies such as virtual reality, augmented reality, and movie special effects, there is a growing demand for high-quality human motion generation. In recent years, several innovative methods and techniques have emerged to tackle this challenging task~\cite{zhu2023human}. Deep generative models, including GANs~\cite{lin2018human,ahn2018text2action}, VAEs~\cite{guo2022generating,petrovich2022temos,petrovich2021action}, and diffusion models~\cite{tevet2023human,zhang2022motiondiffuse,kim2023flame,chen2023executing}, have been widely applied to human motion generation. 

However, there is relatively less research on generating motions in zero-shot settings. In~\cite{tevet2022motionclip}, Tevet et al. proposed MotionCLIP to align human motions with the CLIP space, implicitly injecting the rich semantic knowledge from CLIP into the motion domain to enhance zero-shot generation capability. AvatarCLIP~\cite{hong2022avatarclip} also utilized CLIP to implement a zero-shot text-driven framework for 3D avatar generation and animation. Liang et al.\cite{liang2024omg} pre-trained a large-scale unconditional diffusion model to learn rich out-of-domain motion traits. A Motion ControlNet with a Mixture-of-Controllers block was proposed to align CLIP token embeddings with motion features during fine-tuning.

In order to improve the generalization capability of motion generation models, there have been also attempts to leverage human mesh recovery approaches~\cite{tian2023recovering,zhang2023pymaf,zhang2021pymaf} to collect large-scale pseudo-text-pose datasets~\cite{MAAICCV2023}.
As shown in Azadi et al.~\cite{MAAICCV2023} and Lin et al.~\cite{lin2023being}, the pre-training on such text-pose datasets can improve the generalization to in-the-wild descriptions. However, the static nature of text-pose data makes it difficult to well represent dynamic motions.
Compared to traditional human motion generation, generating motions beyond the distribution of the dataset is more challenging due to the limited scale and diversity of existing motion capture datasets. In Figure~\ref{figCompareFine} and Figure~\ref{figCompareSty}, we show some motions generated by MDM~\cite{tevet2023human} and MLD~\cite{chen2023executing}. Although both approaches achieve impressive performance in human motion generation, they show poor generalization ability since these motions are outside the training data distribution.

Then, with only limited motion capture datasets available, can we still generate motions beyond the distribution of the dataset? For a textual description defining a motion beyond the distribution of the dataset, is there any way to associate it with motions within the dataset? For a never-before-seen motion, the entire body's motion is indeed unseen, but motions of specific body parts might be inside the dataset. Therefore, we can adopt a divide-and-conquer strategy. By re-annotating the motion for different body parts with fine-grained descriptions, we can associate these body parts with specific body parts within the dataset. For example, a vague description ``A person walks depressingly.” can be reformulated as ``His arms hang heavily by his sides. His legs move slowly, taking short steps with little energy...”. Leg movement in this vague unseen motion may appear in ``A person walks aimlessly and slowly.”, of which the motion is included in the dataset. And the arm movement may appear in ``His arms hang heavily by his sides.”, of which the motion is included in the dataset. We give two examples in Figure \ref{figshow}. On one hand, adopting fine-grained descriptions allows the model to understand how the unseen motions are performed in detail. On the other hand, re-annotating motions for different body parts enables the model to learn the essence of motions better. Using fine-grained textual descriptions, we aim to improve the model's zero-shot understanding capability.

Although annotating fine-grained textual descriptions manually for body parts provides more accurate data, it requires significant manual work, resulting in huge costs. Fortunately, with the rapid development of large language models, OpenAI's GPT series models~\cite{ouyang2022training}, known for their outstanding natural language processing capabilities, have gained widespread attention worldwide. In~\cite{gilardi2023chatgpt}, Gilardi et al. demonstrated that ChatGPT performs as well as human annotators in some tasks. In~\cite{Action-GPT}, Action-GPT explores the excellent capability of ChatGPT in expanding human action descriptions. However, the generated content tends to be excessively redundant. For our task, we carefully design a prompt that allows ChatGPT-3.5 to provide detailed but non-redundant transcriptions of text descriptions about human motion. We then use this prompt and ChatGPT-3.5 to transcribe 44,970 short text descriptions from HumanML3D~\cite{guo2022generating} and 6,353 text descriptions from KIT~\cite{plappert2016kit} for model training. We will make these fine-grained transcriptions available to the public.

With these fine-grained descriptions, we propose a new framework named Fine-Grained Human Motion Diffusion Model (FG-MDM) for human motion generation. Specifically, we replace the original simple and vague text with ChatGPT-generated fine-grained descriptions of individual body parts to guide a transformer-based diffusion model. Following MDM~\cite{tevet2023human}, we encode the entire fine-grained description with CLIP~\cite{radford2021learning} as a global token of the transformer. Apart from this global token, we further encode descriptions of different body parts individually with CLIP as part tokens. By adopting these tokens, the model can pay attention to both the global and detailed information of human motions, thereby improving the accuracy and completeness of the denoising results.



\begin{figure*}[t]
\centering
\includegraphics[width=0.99\textwidth]{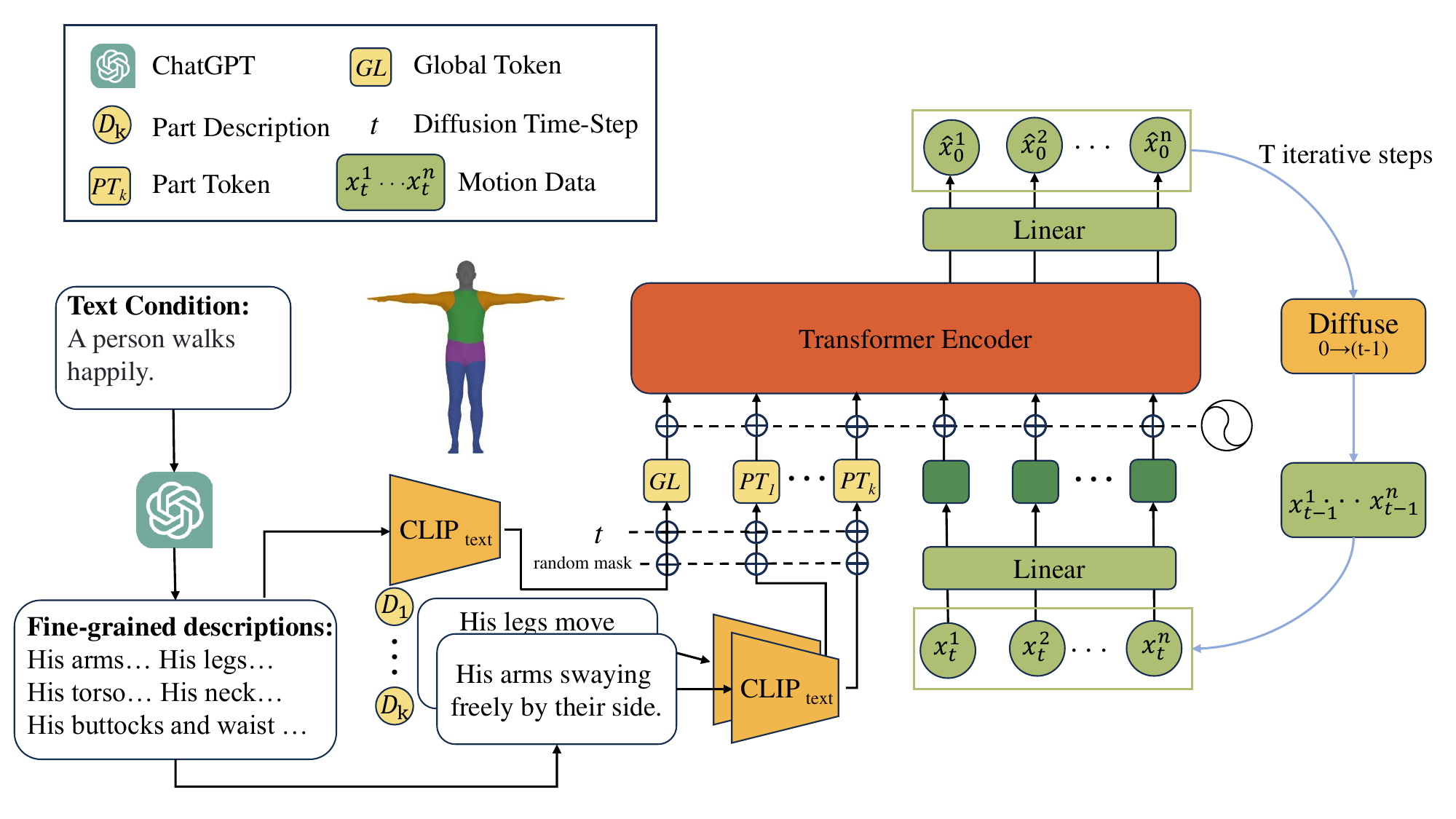} 
\caption{The overall pipeline of FG-MDM. The model learns the denoising process of the diffusion model from the motion $x_{t}^{1:n}$ at time step $t$ to the clean motion $\hat{x}_{0}^{1:n}$, given the text condition. The input text is first paraphrased by ChatGPT into fine-grained descriptions $D_{1:k}$ for different parts of the body, where $k$ denotes the number of body parts. These descriptions are then fed into a pre-trained CLIP text encoder and projected, along with the time step $t$, onto input tokens $PT_{1:k}$ of the transformer. The overall fine-grained text is further encoded into a global input token $GL$, providing holistic information. In the sampling process of the diffusion model, an initial random noise $x_{T}^{1:n}$ is sampled, and then $T$ iterations are performed to generate the clean motion $\hat{x}_{0}^{1:n}$. At each sampling step $t$, guided by $PT_{1:k}$ and $GL$, the transfomer encoder predicts the clean motion $\hat{x}_{0}^{1:n}$ which is then noised back to $x_{t-1}^{1:n}$.}
\label{figframe}
\end{figure*}

Our contributions are summarized as follows:

\begin{itemize}
    \item We present a novel framework that utilizes fine-grained descriptions of different body parts to guide the denoising process of the transformer-based diffusion model. This framework is capable of generating a broader range of motions that extend beyond the distribution of training datasets.
    \item We carefully design a prompt that enables ChatGPT to convert short and vague texts into detailed but non-redundant descriptions of different body parts. We then use this prompt to transcribe 44,970 texts from HumanML3D and 6,353 texts from KIT into fine-grained descriptions. We will make these fine-grained transcriptions publicly available.
    \item We conduct a series of experiments to evaluate our model's ability to not only fit the training data but also generate motions beyond the distribution of the dataset, i.e., the generalization capability. 
\end{itemize}
\section{Related Work}
\label{sec:relatedwork}

\subsection{Human Motion Generation}
There has been a great interest in human motion generation in recent years. Previous work has explored unconditional generative models~\cite{zhao2020bayesian,raab2023modi} as well as generative models using various input conditions, such as text~\cite{guo2022generating,tevet2023human,chen2023executing}, prior motion~\cite{martinez2017human,hernandez2019human}, action class~\cite{guo2020action2motion,petrovich2021action}, and music~\cite{li2021ai,tseng2023edge,li2022danceformer}. In this paper, we focus on text-to-motion generation. Early work usually addressed the text-to-motion task with a sequence-to-sequence model~\cite{lin2018generating}. Later on, the focus shifted beyond simple action labels. For example, Guo et al. utilized variational autoencoders to generate motions from text~\cite{guo2022generating}, significantly enhancing the quality and diversity of generated motions. With the success of diffusion models in AIGC, MDM~\cite{tevet2023human} and other related work~\cite{zhang2022motiondiffuse,kim2023flame,chen2023executing} have introduced diffusion models into the text-to-motion domain, resulting in impressive achievements.

There is a relative scarcity of work that directly focuses on the zero-shot capabilities of motion generation models. However, there has been some work attempting to generate fine-grained or stylized motions to expand the distribution of generated motions. For example, MotionCLIP~\cite{tevet2022motionclip} trained an autoencoder to align motion with the text and images of CLIP in the latent feature space, exploring the generation of unseen motions. MotionDiffuse~\cite{zhang2022motiondiffuse} instead proposed precise control of motions and stylization of body parts using noise interpolation based on the properties of diffusion models. GestureDiffuCLIP~\cite{ao2023gesturediffuclip} incorporated style guidance into the diffusion model through AdaIN~\cite{huang2017arbitrary}, generating realistic and stylized gestures. FineMoGen~\cite{zhang2024finemogen} adopts a novel transformer architecture for generating and editing fine-grained motions. However, there is a significant difference between the methods above and ours. They overlooked that the essence of complex motions described by vague text is the combination of simple movements of body parts. This renders them ineffective when confronted with descriptions of motions they have not encountered.

The work most closely related to ours is Action-GPT~\cite{Action-GPT}, which introduced, for the first time, large language models into the field of text-conditioned motion generation. Action-GPT can be integrated into any text-to-motion model. However, it enriched only the description of action classes without providing detailed descriptions of different body parts and guiding the model training. For another action generation model, SINC~\cite{athanasiou2023sinc} incorporated ChatGPT to identify the body parts involved in the textual description. It achieved impressive results by generating multiple motions and concatenating them using different body parts. Specifically, SINC divides the human body into [‘left arm’, ‘right arm’, ‘left leg’, ‘right leg’, ‘torso’, ‘neck’, ‘buttocks’, ‘waist’], which we borrow from in our work. There is also a costly method that utilizes LLMs. By fine-tuning LLMs, MotionGPT~\cite{jiang2023motiongpt,zhang2023motiongpt} designed a pre-trained motion language model that supports various motion-related tasks through prompts. In contrast, our method is more efficient and can rapidly enhance the model's zero-shot generation capabilities.

\subsection{Diffusion Generative Models} 
The diffusion model is a neural generative model based on the stochastic diffusion process in thermodynamics~\cite{sohl2015deep,ho2020denoising}. It starts with samples from the data distribution and gradually adds noise through a forward diffusion process. Then, a neural network learns the reverse process to progressively remove the noise and restore the samples to their original states. Diffusion generative models have achieved significant success in the image generation field~\cite{ramesh2022hierarchical,rombach2022high}. For conditional generation, ~\cite{dhariwal2021diffusion} introduced classifier-guided diffusion, while ~\cite{ho2022classifier} proposed a classifier-free method. Given their excellent generation quality, \cite{tevet2023human,zhang2022motiondiffuse,kim2023flame} incorporated diffusion models into the motion generation domain, leading to impressive results.
\section{Method}
\label{sec:method}

Given a textual description, our goal is to generate a human motion $x^{1:n}=\{x^{i}\}_{i=1}^{n}$ that matches the given description. The motion consists of $n$ frames of human poses. For each pose $x^{i}\in\mathbb{R}^{J\times D}$, we represent it by joint rotations or positions, where $J$ represents the number of joints and $D$ represents the dimensionality of the joint representation. In Figure \ref{figframe}, we give an overview of our fine-grained human motion diffusion model. First, we adopt ChatGPT to perform fine-grained paraphrasing of the vague textual description. This expands concise textual descriptions into descriptions of different body parts. FG-MDM then uses these fine-grained descriptions to guide a diffusion model for human motion generation.

\subsection{Prompt Strategy}
\label{sec:prompt}
We first introduce the prompt strategy adopted for generating fine-grained descriptions. We utilize ChatGPT-3.5 to create more fine-grained descriptions based on different body parts for a given textual description of a motion. ChatGPT is a conversational model based on a large language model that can engage in natural conversations and generate corresponding responses. The answers from ChatGPT are often directly influenced by the information and expression provided in the prompt. If the prompt offers clear and detailed questions or instructions, ChatGPT can typically provide relevant and accurate answers. However, if the prompt is too simple, ambiguous, or unclear, ChatGPT may generate unexpected responses or express unclear content. For our task, we carefully design an effective prompt by using experimental verification.

Our designed prompt is: ``Translate the motion described by the given sentences to the motion of each body part only using one paragraph. The available body parts include [`arms', `legs', `torso', `neck', `buttocks', `waist’]. Here are some examples:[Q...A...]. Question: [sentence]”. [sentence] is the vague textual description that needs to be refined. [Q...A...] are four examples of Q\&A pairs designed manually. Here is one of the examples,

\textbf{Question}: A man kicks something or someone with his left leg.

\textbf{Answer}: His left leg extends out with force as he kicks something or someone. His arms are held steady at his sides. His torso is slightly twisted. His buttocks and waist muscles are contracted. His neck remains neutral.

These examples are essential as they are critical in aiding LLMs in maintaining the structural stability of their output results. This design guarantees the stable quality of the fine-grained descriptions while re-annotating thousands of textual descriptions from HumanML3D and KIT.

\subsection{Diffusion Model for Motion Generation}
The basic idea of diffusion models~\cite{sohl2015deep,ho2020denoising} is to learn the reverse process of a well-defined stochastic process. Following MDM~\cite{tevet2023human}, we design a text-driven human motion generation model based on the diffusion model.

The diffusion model consists of the forward process and the reverse process, both of which follow the Markov chain. The forward process involves adding noise. The input is the original motion $x_{0}^{1:n}$ from the data distribution, and the output is the motion $x_{t}^{1:n}$ with adding Gaussian noise $t$ times. When enough noise is added, the motion $x_{T}^{1:n}$ can approach the Gaussian distribution $\mathcal{N}(\mathbf{0}, \mathbf{I})$. The reverse process aims to reduce the noise in the Gaussian noise $x_{T}^{1:n} \sim \mathcal{N}(\mathbf{0}, \mathbf{I})$. In the denoising process, at diffusion step $t$, a portion of the noise is eliminated, resulting in a less noisy motion $x_{t-1}^{1:n}$. This step is repeated iteratively until the noise is completely removed, generating a clean motion $\hat{x}_{0}^{1:n}$.

\subsubsection{Network}
we adopt a simple transformer~\cite{vaswani2017attention} encoder architecture to implement our network $G$. Unlike the conventional diffusion model mentioned above, we follow~\cite{ramesh2022hierarchical} and predict the clean motion $\hat{x}_{0}^{1:n}$ instead of predicting the noise added in each time-step. The input of $G$ is the noised motion $x_{t}^{1:n}$ obtained by adding noise $t$ times to the original motion $x_{0}^{1:n}$. The noised motion $x_{t}^{1:n}$, together with the text condition tokens $GL$, $PT_{1:k}$ and the time-step $t$, are inputted to the transformer encoder, resulting in the clean motion $\hat{x}_{0}^{1:n}$. One of the reasons for directly predicting the clean motion in each time-step of the diffusion model is to incorporate human geometric losses during the training of the network, making the generated human motions more natural. For each sampling step $t$, from $T$ to 1, our model predicts the clean motion $\hat{x}_{0}^{1:n}$, and then adds noise back to $x_{t-1}^{1:n}$. After $T$ iterations, the final clean motion $\hat{x}_{0}^{1:n}$ is obtained. This form of diffusion model has become commonly adopted, as do we.

\subsubsection{Global Token and Part Tokens} For the text condition, we encode the entire fine-grained description with CLIP~\cite{radford2021learning} as a global token $GL$ of the transformer. Apart from this global token, we further encode descriptions of different body parts individually with CLIP as part tokens $PT_{1:k}$, where $k$ denotes the number of body parts. The global token serves as an overall condition to guide the diffusion process. Part tokens provide explicit information for fine-grained control of the movements of each body part. Part tokens effectively make up for the ambiguity of the original description text. It greatly enhances our FG-MDM's ability to understand in-the-wild text, making it outstanding on zero-shot generation tasks.


\subsubsection{Loss Functions} For training the diffusion model, we follow ~\cite{ramesh2022hierarchical} to predict the signal itself instead of predicting the noise, i.e., $\hat{x}_{0}^{1:n}\,=\,G(x_{t}^{1:n},t,c)$, with the simple loss function.
\begin{equation}
\mathcal{L}_{\mathrm{G}}=E_{x_{0}^{1:n}\sim q(x_{0}^{1:n}|c),t\sim[1,T]}[||x_{0}^{1:n}-G(x_{t}^{1:n},t,c)]|^{2}]
\end{equation}

In order to generate more natural and kinematically plausible motions, we employ the same geometric losses as MDM~\cite{tevet2023human} from~\cite{shi2020motionet,petrovich2021action}, i.e., positions, foot contact, and velocities.
\begin{equation}
\mathcal{L}_{\mathrm{pos}}=\frac{1}{n}\sum_{i=1}^{n}||F K(x_{0}^{i})-F K(\hat{x}_{0}^{i})||_{2}^{2},
\end{equation}
\begin{equation}
\mathcal{L}_{\mathrm{foot}}=\frac{1}{n-1}\sum_{i=1}^{n-1}||(F K(\hat{x}_{0}^{i+1})-F K(\hat{x}_{0}^{i}))\cdot f_{i}||_{2}^{2},
\end{equation}
\begin{equation}
\mathcal{L}_{\mathrm{vel}}=\frac{1}{n-1}\sum_{i=1}^{n-1}||(x_{0}^{i+1}-x_{0}^{i})-(\hat{x}_{0}^{i+1}-\hat{x}_{0}^{i})||_{2}^{2}
\end{equation}
where $F K(\cdot)$ represents the forward kinematic function that converts joint rotations into joint positions. For each frame $i$, $f_{i}\in\{0,1\}^{J}$ is the binary foot contact mask.

Overall, our training loss is
\begin{equation}
\mathcal{L}=\mathcal{L}_{\mathrm{G}}+\lambda_{\mathrm{pos}}\mathcal{L}_{\mathrm{pos}}+\lambda_{\mathrm{vel}}\mathcal{L}_{\mathrm{vel}}+\lambda_{\mathrm{foot}}\mathcal{L}_{\mathrm{foot}}.
\end{equation}
where $\lambda_{\mathrm{pos}}$, $\lambda_{\mathrm{vel}}$, $\lambda_{\mathrm{foot}}$ are balancing coefficients for the three geometric losses.
\section{Experiments}
\label{sec:experiments}

\begin{figure*}[t]
    \centering
    \includegraphics[width=0.99\textwidth]{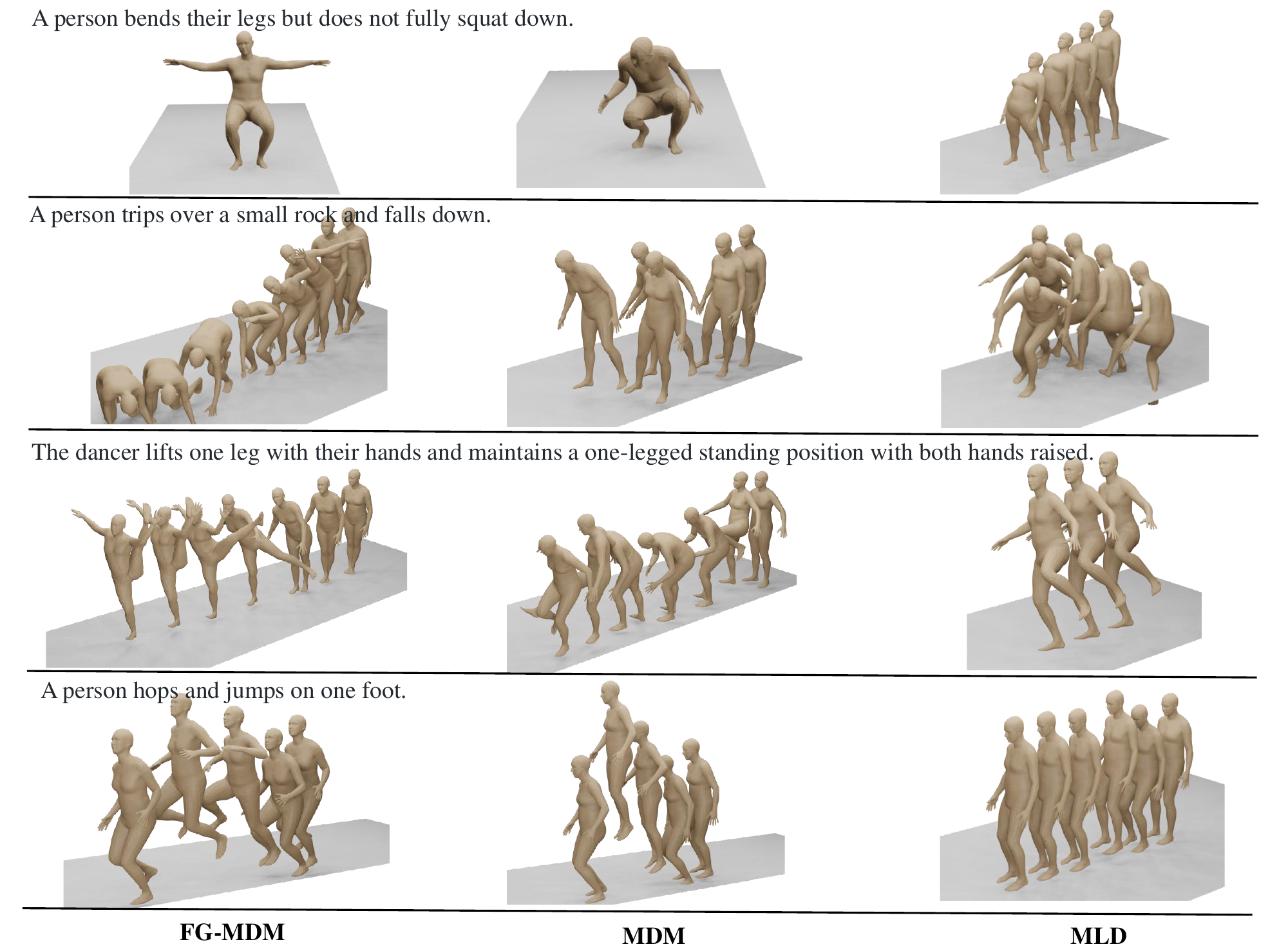} 
    \caption{Qualitative results with unseen motions. We compare our FG-MDM with MDM~\cite{tevet2023human} and MLD~\cite{chen2023executing}. All three models are trained on HumanML3D. For better visualization, some pose frames are shifted to prevent overlap. Please refer to supplementary materials for more video demos.}
    \label{figCompareFine}
\end{figure*}

\begin{figure*}[t]
    \centering
    \includegraphics[width=0.99\textwidth]{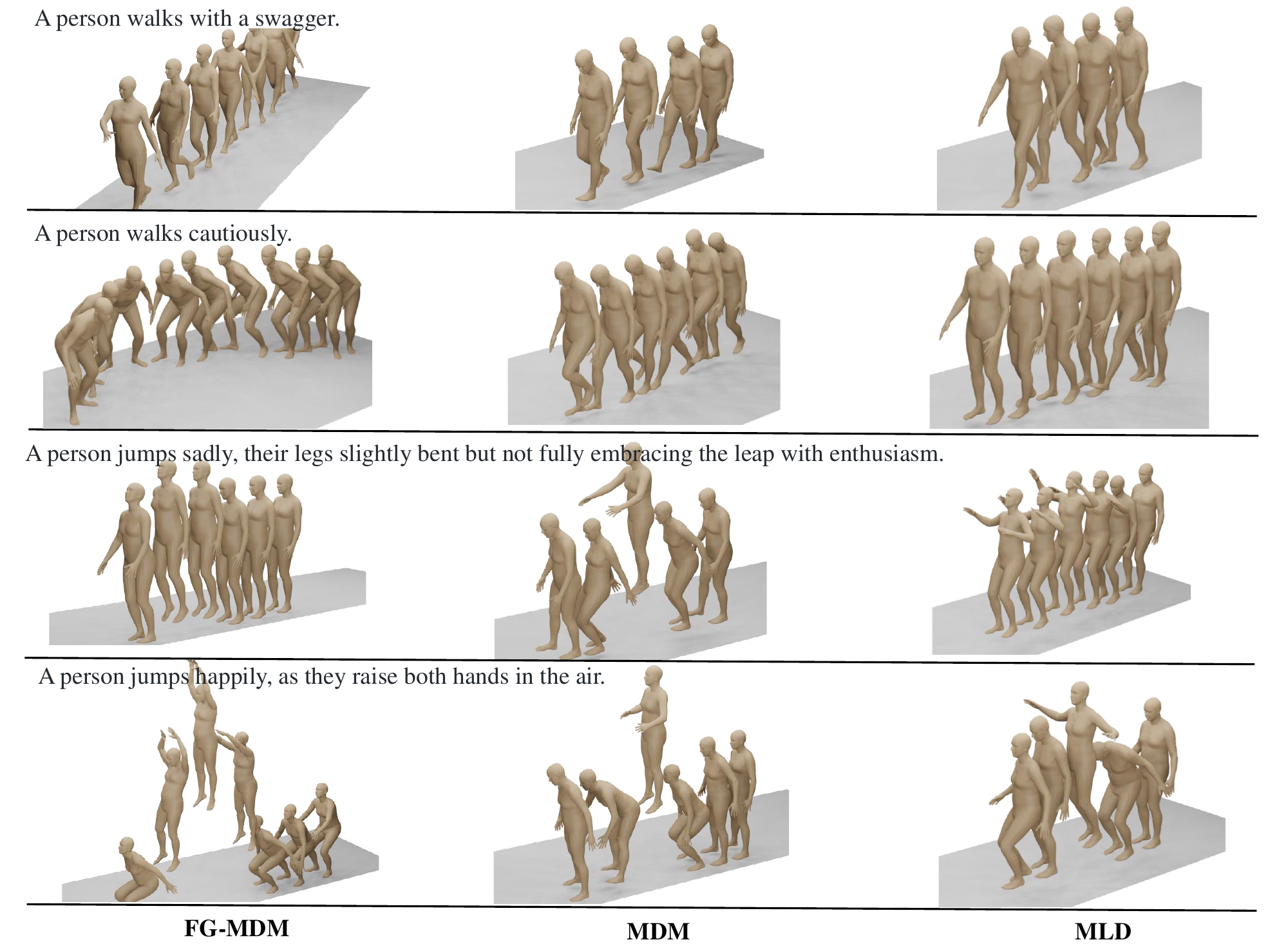} 
    \caption{Qualitative results with unseen stylized motions. All three models are trained on HumanML3D. Please refer to supplementary materials for more video demos.}
    \label{figCompareSty}
\end{figure*}

\begin{table*}[t]
\caption{Quantitative results on HumanML3D and HuMMan. The model marked with * indicates that both the ChatGPT-Refined text and the manually annotated text provided by HumanML3D are used during training. \textbf{Bold text} is the best result, \underline{underlined text} is the second-best result. Zero-shot means that the models are evaluated directly on HuMMan after training on HumanML3D.}
\label{tab1}
\centering
\resizebox{0.9\textwidth}{!}{
\begin{tabular}{lccc|ccc}
\toprule
\multirow{2}{*}{Methods} & \multicolumn{3}{c}{HumanML3D} & \multicolumn{3}{c}{HuMMan(zero-shot)} \\
\cmidrule(lr){2-4} \cmidrule(lr){5-7}
& FID$\downarrow$ & MM Dist$\downarrow$ & Diversity$\uparrow$ & FID$\downarrow$ & MM Dist$\downarrow$ & Diversity$\uparrow$ \\

\midrule
Real & $0.002^{\pm.000}$ & $2.974^{\pm.008}$ & $9.503^{\pm.065}$ & $0.032^{\pm.002}$ &$23.019^{\pm.042}$ & $4.709^{\pm.097}$ \\
\midrule
MAA~\cite{MAAICCV2023} & $0.774^{\pm.007}$ & - & $8.230^{\pm.064}$ & - & - & - \\
T2M-GPT~\cite{zhang2023generating} & $\textbf{0.116}^{\pm.004}$ & $\underline{3.118}^{\pm.011}$ & $\textbf{9.761}^{\pm.081}$ & $\textbf{9.631}^{\pm.203}$ & $27.582^{\pm.073}$ & $5.149^{\pm.145}$ \\
MLD~\cite{chen2023executing}  & $0.473^{\pm.013}$ & $3.196^{\pm.010}$ & $\underline{9.724}^{\pm.082}$ & $14.970^{\pm.472}$ &$27.104^{\pm.024}$ & $5.493^{\pm.101}$ \\
MotionDiffuse~\cite{zhang2022motiondiffuse}  & $0.630^{\pm.001}$ & $\textbf{3.113}^{\pm.001}$ & $9.410^{\pm.049}$ & $30.138^{\pm.712}$ &$28.747^{\pm.041}$ & $5.357^{\pm.045}$ \\
MDM~\cite{tevet2023human} & $0.544^{\pm.044}$ & $5.566^{\pm.027}$ & $9.559^{\pm.086}$ & $13.375^{\pm.408}$ &$27.689^{\pm.055}$ & $5.585^{\pm.089}$ \\
\midrule
FG-MDM & $0.663^{\pm.012}$ & $5.649^{\pm.024}$ & $9.476^{\pm.068}$ & $17.180^{\pm.272}$ &$\underline{26.867}^{\pm.030}$ & $\underline{5.589}^{\pm.124}$ \\
FG-MDM* & $0.618^{\pm.009}$ & $5.274^{\pm.048}$ & $9.563^{\pm.097}$ & $\underline{12.460}^{\pm.330}$ &$\textbf{26.814}^{\pm.019}$ & $\textbf{5.626}^{\pm.100}$ \\
\bottomrule
\end{tabular}
}
\end{table*}
\begin{table}[]
\caption{Quantitative results on Kungfu. Zero-shot means that the models are evaluated directly on Kungfu after training on HumanML3D.}
\label{tab2}
\centering
\resizebox{.99\columnwidth}{!}{
\begin{tabular}{lccc}
\toprule
\multirow{2}{*}{Methods}  & \multicolumn{3}{c}{Kungfu(zero-shot)} \\
\cmidrule(lr){2-4} 
& FID$\downarrow$ & MM Dist$\downarrow$ & Diversity$\uparrow$ \\

\midrule
Real & $0.133^{\pm.010}$ & $22.164^{\pm.041}$ & $5.351^{\pm.312}$ \\
\midrule
MAA~\cite{MAAICCV2023} & - & - & -\\
T2M-GPT~\cite{zhang2023generating}  & $\textbf{12.652}^{\pm.429}$ & $\underline{25.826}^{\pm.041}$ & $5.702^{\pm.428}$\\
MLD~\cite{chen2023executing}  & $18.524^{\pm.352}$ & $27.182^{\pm.020}$ & $5.598^{\pm.356}$\\
MotionDiffuse~\cite{zhang2022motiondiffuse}  & $26.363^{\pm.337}$ & $26.320^{\pm.035}$ & $\textbf{6.117}^{\pm.691}$\\
MDM~\cite{tevet2023human}  & $16.396^{\pm.466}$ & $26.280^{\pm.095}$ & $5.468^{\pm.590}$\\
\midrule
FG-MDM  & $19.340^{\pm.797}$ & $26.845^{\pm.052}$ & $5.142^{\pm.759}$\\
FG-MDM*  & $\underline{15.892}^{\pm.567}$ & $\textbf{25.325}^{\pm.035}$ & $\underline{5.814}^{\pm.479}$\\
\bottomrule
\end{tabular}
}
\end{table}

In this section, we first elaborate the datasets, evaluation metrics, and implementation details in Section~\ref{sec:details}. We then conduct quantitative experiments to compare FG-MDM with current state-of-the-art approaches in Section~\ref{sec:prior}. To show the generalization capability of our model, we further perform quantitative experiments, qualitative experiments, and a user study to examine FG-MDM's ability to generate motions beyond the distribution of training datasets. To evaluate our method comprehensively, we design two additional ablation experiments in Section~\ref{sec:ablation}. The process of designing prompts will also be elaborated in this section.

\subsection{Experimental Details}
\label{sec:details}

\subsubsection{Datasets} 
We utilize the HumanML3D~\cite{guo2022generating} dataset and the KIT~\cite{plappert2016kit} dataset to train and evaluate our model. The HuMMan~\cite{cai2022humman} dataset and Kungfu dataset from the Motion-X dataset~\cite{lin2023motionx, humantomato} are employed to assess the models' zero-shot performance. HumanML3D is a recently proposed large-scale dataset of motion-text pairs. It consists of 14,616 motion sequences from the AMASS~\cite{mahmood2019amass} and HumanAct12~\cite{guo2020action2motion} datasets, with multiple ways of describing each motion, resulting in a total of 44,970 text annotations. The KIT dataset, on the other hand, is relatively smaller and contains 3,911 motion sequences along with their corresponding 6,353 text descriptions. For both datasets, we use 80\% of the data for training and the remaining for testing. Motion-X is a large-scale dataset of whole-body motions and whole-body pose annotations, integrating several existing datasets and additional online videos. For zero-shot testing, we utilize 100\% of the HuMMan and Kungfu subsets from it. HuMMan is a multi-modal human dataset, containing 744 motion sequences and their corresponding 744 texts descriptions. Kungfu encompasses many human motions related to martial arts, with a total of 1040 motion sequences and their corresponding 1040 texts descriptions.

We preprocess the 44,970 text descriptions from HumanML3D and 6,353 text descriptions from KIT using ChatGPT-3.5. This preprocessing extends these descriptions into fine-grained ones for our model training.

\subsubsection{Evaluation Metrics} 
We employ three evaluation metrics for quantitative experiments to evaluate our model's ability to fit the training data: FID, Multimodal Dist, and Diversity. Multimodal Dist assesses the correlation between generated motions and input text. Diversity is utilized to evaluate the diversity of generated motions. FID measures the difference in feature distribution between generated motions and ground truth in latent space, which is used to evaluate the quality of generated motions.

\subsubsection{Implementation Details}
In our study, the transformer accepts tokens whose feature dimension is 512 as input. We use four attention heads and apply a dropout rate of 0.1. The transformer encoder consists of 8 stacked encoder layers to capture complex relationships and hierarchies in the data. For ChatGPT, we adopt the gpt-3.5-turbo API provided by OpenAI. For text encoding, we employ the frozen CLIP-ViT-B/32 model as the encoder. Our batch size is set to 64. Additionally, we set the diffusion step to 1000. On a single NVIDIA GeForce RTX3090 GPU, it takes about six days to train our model.

\subsection{Comparison with Prior Work} 
\label{sec:prior}

To evaluate the performance of FG-MDM in handling zero-shot text-conditioned motion generation, we compare our work with five recent motion generation approaches: MAA~\cite{MAAICCV2023}, T2M-GPT~\cite{zhang2023generating}, MLD~\cite{chen2023executing}, MotionDiffuse~\cite{zhang2022motiondiffuse}, and MDM~\cite{tevet2023human}. In Table \ref{tab1} and Table \ref{tab2}, we provide experimental results on the HumanML3D, HuMMan, and Kungfu datasets, respectively. For all experiments, We run the evaluation five times, and ``±'' indicates the 95\% confidence interval. For the six SOTA methods, on HumanML3D, we directly cite their results reported in their original papers. To examine the generalization ability of the methods, we use HumanML3D as the training set and HuMMan and Kungfu as the test sets. To do so, we train TMR~\cite{petrovich23tmr,petrovich22temos} using the HuMMan and Kungfu datasets to obtain a pair of text encoder and motion encoder for calculating the MM Dist metric.
For SOTA methods, we apply their released pre-trained models on HumanML3D to HuMMan and Kungfu to evaluate their zero-shot generation performance. Since MAA~\cite{MAAICCV2023} does not release the pre-trained model, we cannot test its zero-shot generation performance.

When evaluated on the test set of HumanML3D, all five methods achieve state-of-the-art performance. For FG-MDM, the ChatGPT-Refined fine-grained textual description doesn't match the manually annotated textual description better. Therefore, under within-dataset settings, our model does not exceed those SOTA models on HumanML3D. However, on the HuMMan and Kungfu datasets, FG-MDM captures most of the best and second-best results. Note that the size of our training dataset is much smaller than some SOTA methods like \cite{MAAICCV2023}, but we still demonstrate solid zero-shot capabilities.


Existing quantitative results do not reflect the real generation performance of the model well. Therefore, we provide some qualitative results to let readers intuitively feel the superiority of our method. In Figure~\ref{figCompareFine} and Figure~\ref{figCompareSty}, we show motions generated by MDM~\cite{tevet2023human}, MLD~\cite{chen2023executing} and our FG-MDM. Note that for all three methods, we use models trained on HumanML3D to generate motions. In comparison, our method generates motions more consistent with the details described in the fine-grained textual descriptions. This shows that our divide-and-conquer method works. Motion generation models require clear and specific conditions to generate the motions needed. Even for difficult stylized descriptions, our method can generate motions that correspond to the style in the textual descriptions. This demonstrates the strong generalization capability of FG-MDM. Please refer to supplementary materials for more video demos.

\begin{table}[]
\caption{Ablation study results on HumanML3D. ``Fine-grained" denotes using ChatGPT-generated fine-grained descriptions. ``Part" represents adopting part tokens. Note that the models are trained on HumanML3D.}
\label{tab3}
\centering
\resizebox{\columnwidth}{!}{
\begin{tabular}{ccccc}
\toprule
Fine-grained  & Part  & FID$\downarrow$  & MM Dist$\downarrow$ & Diversity$\uparrow$ \\ 
\midrule
          &     & $4.363$            &  $7.298$         & $8.432$   \\
    \checkmark  &     & $1.050$            &  $6.778$         & $\textbf{9.509}$   \\
\midrule
    \checkmark  &   \checkmark   & $\textbf{0.663}$   &  $\textbf{5.649}$   & $9.476$   \\
\bottomrule
\end{tabular}
}
\end{table}

\begin{table}[]
\caption{Ablation study results on KIT. Note that the models are trained on KIT.}
\label{tab4}
\centering
\resizebox{\columnwidth}{!}{
\begin{tabular}{ccccc}
\toprule
Fine-grained  & Part & FID$\downarrow$  & MM Dist$\downarrow$ & Diversity$\uparrow$ \\ 
\midrule
          &     & $16.372$            &  $10.502$        & $8.758$   \\
    \checkmark  &     & $0.549$            &  $9.826$         & $\textbf{10.829}$   \\
\midrule
    \checkmark  &   \checkmark   & $\textbf{0.344}$   &  $\textbf{9.352}$   & $10.707$   \\
\bottomrule
\end{tabular}
}
\end{table}

\subsection{Ablation Study}
\label{sec:ablation}

To validate our contribution, we conduct two ablation studies. As shown in Table \ref{tab3} and Table \ref{tab4}, the first row shows our baseline. The first study examines the contribution of ChatGPT-Generated fine-grained texts, which is performed by replacing the original short text with the fine-grained description. The improvement can be said to be huge. We cleverly utilize the powerful reasoning capabilities of LLMs and let them help our generative model better understand the nature of text conditions, bringing a leap to the zero-shot performance. The second study checks the contribution of part tokens when fine-grained descriptions are used. As observed, a reasonable framework also improves the quality of generated motions. However, perhaps more conditions bring more constraints to generations, leading to a decrease in diversity. But this drop is acceptable. So, we finally adopt the design of part tokens. 

The process of designing a prompt is also worth introducing. We put a lot of effort into prompt engineering to better utilize LLMs. Initially, we wanted to replace the abstract emotion words in the text with concrete motion descriptions. The prompt is designed as “Replace the emotional words in this sentence with corresponding descriptions of human body movements, reorganize the language, and output new sentences. [sentence]”. However, GPT cannot understand the task requirements well, and the answers are incorrect. For example, the answers describe muscles, facial expressions, etc., and have nothing to do with human skeletal movements. Then we follow Action-GPT~\cite{Action-GPT} to make the prompt as “Describe a person's body movements who is performing the action [sentence] in details”. But the output is not concise enough. There would be redundant information and irrelevant information. The structure of the output content is also irregular. For example, The granularity is too fine, and some sentences have no action information. So we decided to divide the body parts to make the output cleaner, i.e., “Use a concise and clear paragraph to describe the body movements of the person executing [sentence] using one paragraph. body movements only includes  [‘arms’, ‘legs’, ‘torso’, ‘neck’, ‘buttocks’, ‘waist’].” But it still cannot completely solve this problem. Finally, we add in-context learning to improve output quality. By letting LLMs learn from the examples we give, its output content is standardized, making it better meet our needs. The final prompt is shown in Section~\ref{sec:prompt}.

\begin{figure}[h]
\centering
\includegraphics[width=0.99\columnwidth]{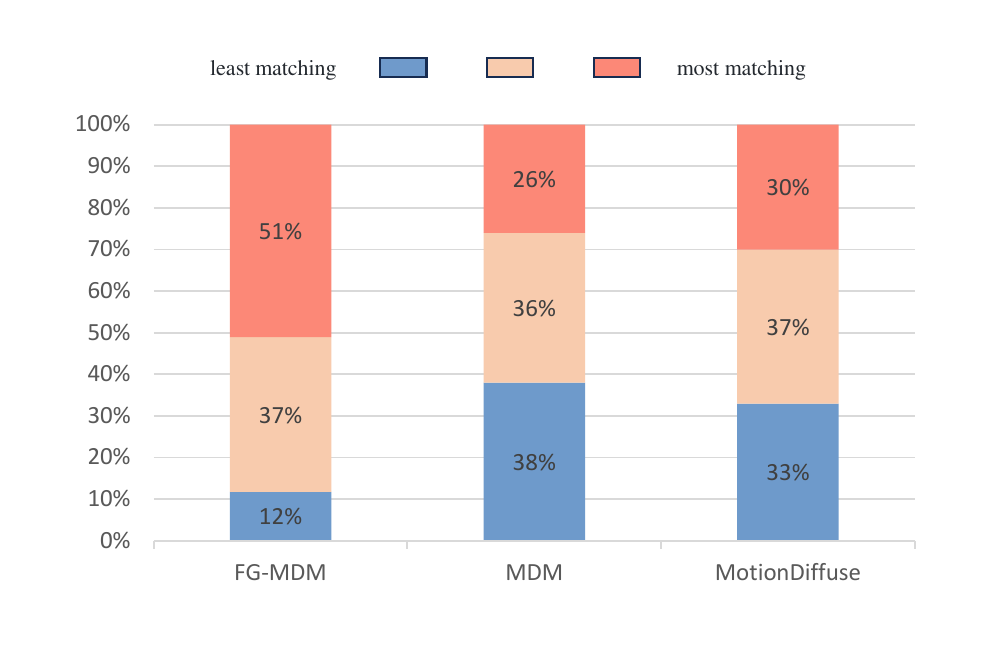} 
\caption{User study results. For each method, a color bar ranging from blue to red represents the percentage of text-to-motion match levels, with blue indicating the least match and red indicating the most match.
}
\label{figuser}
\end{figure}

\subsection{User Study} 

To further examine FG-MDM's generalization capability, we conduct a user study to evaluate the quality of motions generated by our model based on human visual perception. We customize a total of 40 textual descriptions beyond the distribution of the dataset. With these descriptions, we generate motions by using MDM~\cite{tevet2023human}, MotionDiffuse~\cite{zhang2022motiondiffuse}, and our FG-MDM. We then recruit ten users for the study. In each question, participants are asked to rate the degree of matching between the generated motion and the textual description on a scale of 0 to 2. The results are given in Figure \ref{figuser}. Apparently, FG-MDM matches texts much better in generating motions beyond the distribution of the dataset than the other two methods. Nearly half of the generated motions get the highest score. In contrast, MDM and MotionDiffuse perform poorly. Most of the generated motions are not satisfactory.



\section{Conclusion}
In this study, we used LLMs to perform fine-grained paraphrasing on the textual annotations of HumanML3D and KIT. With these fine-grained descriptions, We explored a Fine-Grained Human Motion Diffusion Model. It utilizes fine-grained descriptions of different body parts to guide the training of a diffusion model. This enables it to learn the essence of motions and thus generate motions beyond the distribution of training datasets. In the future, we would like to improve the quality of fine-grained annotations of human motions. Having high-quality text labels will greatly promote research on human motion generation.
{
    \small
    \bibliographystyle{ieeenat_fullname}

}


\end{document}